\documentclass[,nonblindrev]{informs3} 

\OneAndAHalfSpacedXII 

\usepackage{natbib}
 \bibpunct[, ]{(}{)}{,}{a}{}{,}%
 \def\bibfont{\small}%

\usepackage{amsmath}
\usepackage{amsfonts}
\usepackage{amssymb}
\usepackage{amsbsy}
\usepackage{subfigure}
\usepackage{latexsym}
\usepackage{graphicx}
\usepackage{mathptmx}
\DeclareMathAlphabet{\mathcal}{OMS}{cmsy}{m}{n}
\DeclareSymbolFont{largesymbols}{OMX}{cmex}{m}{n}

\usepackage{diagbox}
\usepackage{dsfont}
\usepackage{algorithmicx}
\usepackage{algorithm}
\usepackage{algpseudocode}

\usepackage{booktabs}

\makeatletter
\newenvironment{breakablealgorithm}
  {
   \begin{center}
     \refstepcounter{algorithm}
     \hrule height.8pt depth0pt \kern2pt
     \renewcommand{\caption}[2][\relax]{
       {\raggedright\textbf{\ALG@name~\thealgorithm} ##2\par}%
       \ifx\relax##1\relax 
         \addcontentsline{loa}{algorithm}{\protect\numberline{\thealgorithm}##2}%
       \else 
         \addcontentsline{loa}{algorithm}{\protect\numberline{\thealgorithm}##1}%
       \fi
       \kern2pt\hrule\kern2pt
     }
  }{
     \kern2pt\hrule\relax
   \end{center}
  }
\makeatother

\usepackage{epstopdf}

\usepackage{bbm}

\usepackage{array}
\usepackage{tikz}
\newcommand*{\circled}[1]{\lower.7ex\hbox{\tikz\draw (0pt, 0pt)%
    circle (.5em) node {\makebox[0.0em][c]{\small #1}};}}

\newcolumntype{P}[1]{>{\centering\arraybackslash}p{#1}}
\newcolumntype{M}[1]{>{\centering\arraybackslash}m{#1}}

\EquationsNumberedThrough    
\newtheorem{prop}{Proposition}

\usepackage{color}

\begin{document}
\RUNAUTHOR{Zhang, Peng and Xu}

\RUNTITLE{Efficient Sampling Policy For MCTS}

\TITLE{An Efficient Dynamic Sampling Policy For Monte Carlo Tree Search}
\ARTICLEAUTHORS{%
\AUTHOR{Gongbo Zhang}
\AFF{Department of Management Science and Information Systems, Guanghua School of Management
	Peking University, Beijing, 100871 CHINA.
	\EMAIL{gongbozhang@pku.edu.cn}}
\AUTHOR{Yijie Peng}
\AFF{Department of Management Science and Information Systems, Guanghua School of Management
	Peking University, Beijing, 100871 CHINA.
 \EMAIL{pengyijie@pku.edu.cn}} 
\AUTHOR{Yilong Xu}
\AFF{Department of Computer Science, Beijing Jiaotong University, Beijing, 100044 CHINA.
 \EMAIL{18281110@bjtu.edu.cn}}
}
	
\ABSTRACT{%
We consider the popular tree-based search strategy within the framework of reinforcement learning, the Monte Carlo Tree Search (MCTS), in the context of finite-horizon Markov decision process. We propose a dynamic sampling tree policy that efficiently allocates limited computational budget to maximize the probability of correct selection of the best action at the root node of the tree. Experimental results on Tic-Tac-Toe and Gomoku show that the proposed tree policy is more efficient than other competing methods.}


\maketitle

\section{INTRODUCTION}\label{sec:intro}

Monte Carlo Tree Search (MCTS) is a popular tree-based search strategy within the framework of reinforcement learning (RL), which estimates the optimal value of a state and action by building a tree with Monte Carlo simulation. It has been widely used in sequential decision makings, including scheduling problems, inventory, production management, and real-world games, such as Go, Chess, Tic-tac-toe and Chinese Checkers. See~\cite{browne2012survey,fu2018monte} and \cite{swiechowski2021monte} for thorough overviews. MCTS uses little or no domain knowledge and self learns by running more simulations. Many variations have been proposed for MCTS to improve its performance. In particular, deep neural networks are combined into MCTS to achieve a remarkable success in the game of Go \citep{silver2016mastering,silver2017mastering}.

A basic MCTS is to build a game tree from the root node in an incremental and asymmetric manner, where nodes correspond to states and edges correspond to possible state-action pairs. For each round of MCTS, a tree policy is used to find a node from which a roll-out (simulation) is then performed, and nodes in the collected search path is updated according to the received terminal reward. Moves are made during the roll-out by a default policy, which in the simplest case is to make uniform random moves. Different from depth-limited minimax search that needs to evaluate values of intermediate states, only the reward of the terminal state at the end of each roll-out is evaluated in MCTS, which greatly reduces the amount of domain knowledge required. The best action of the root node is selected based on the information collected from simulations after computational budget is exhausted. The tree policy plays a vital role in the success of MCTS since it determines how the tree is built and computational budget is allocated in simulations. The key issue is to balance the exploration of nodes that have not been well sampled yet and the exploitation of nodes that appear to be promising. In our work, we propose a new tree policy to improve the performance of MCTS.

One of the popular tree policies in MCTS is the Upper Confidence Bounds for Trees (UCT) algorithm, which is proposed by applying the Upper Confidence Bound (UCB1) algorithm \citep{auer2002finite}---originally designed for stochastic multi-armed bandit (MAB) problems---to each node of the tree \citep{kocsis2006bandit,kocsis2006improved}. Stochastic MAB is a well-known sequential decision problem in which the goal is to maximize the expected total reward in finite rounds by choosing amongst finitely many actions (also known as arms of slot machines in the MAB literature) to sample. There are other variants of bandit-based methods developed for the tree policy. \cite{auer2002finite} introduce UCB1-Tuned in order to tune the bounds of UCB1 more finely. \cite{tesauro2012bayesian} suggest a Bayesian framework inspired by its more accurate estimation of values and uncertainties of nodes under limited computational budget. \cite{teytaud2011upper} employ the Exploration-Exploitation with Exponential weights in conjunction with UCT to deal with partially observable games with simultaneous moves. \cite{mansley2011sample} combine the Hierarchical Optimistic Optimisation into the roll-out planning, overcoming the limitation of UCT for a continuous decision space. \cite{teraoka2014efficient} propose a tree policy by selecting the node with the largest confidence interval inspired by the Best Arm Identification (BAI) problem in the MAB literature \citep{bubeck2012regret}, and \cite{kaufmann2017monte} further extend their results to a tighter upper bound. However, both tree policies are pure exploration policies and only developed for the min-max game trees.


Although the goal in MCTS is very similar to the MAB problem, i.e., choosing an action at given state with the best average reward, their setups have many differences. Stochastic rewards are collected at all rounds in MAB, whereas in MCTS, the reward of the goal is collected only at the end of the algorithm. Most bandit-based methods assume that rewards are bounded and known---typically assumed to be $\left[0,1\right]$---however, a more general tree search problem has an unknown and unbounded range of values of nodes. A common objective function of bandit-based methods is the cumulative regret, i.e., the expected sum of difference between the performance of the best arm and that of the chosen arm for sampling. \cite{li2021optimal} show that the algorithms designed to minimize regret tend to discourage exploration. In addition to the differences mentioned above, most bandit-based tree policies only consider the average value and the number of visits of nodes, which do not utilize other available information such as variances. These findings lead us to formulate the tree policy as a statistical ranking and selection (R\&S) problem \citep{chen2011stochastic,powell2012ranking} that has been actively studied in simulation optimization. In statistical R\&S, the goal is to efficiently allocate limited computational budget to finitely many actions (also known as alternatives in the R\&S literature) so that the probability of correct selection (PCS) for the best action can be maximized. The samples for any action are usually assumed to be independent and identically Gaussian distributed with known variances, and a reward is collected after computational budget is exhausted. Despite the same goal of BAI and R\&S, different assumptions on distributions of samples are made. In particular, the former assumes samples to be bounded or sub-Gaussian distributed.

In our work, we aim to maximize the PCS for the optimal action at the root node of the tree. We propose a dynamic sampling tree policy by applying the Asymptotically Optimal Allocation Policy (AOAP) algorithm \citep{peng2018ranking}, which is originally designed for statistical R\&S problems. AOAP is a myopic sampling procedure that maximizes a value function approximation one-step look ahead. The closest work to our paper is \cite{li2021optimal}, where they propose a tree policy by applying the Optimal Computing Budget Allocation (OCBA) algorithm \citep{chen2000simulation,chen2011stochastic} to each node of the tree. The key algorithmic differences from ours lie in: OCBA is developed based on a static optimization problem and is designed to reach a good asymptotic
behavior, whereas AOAP is derived in a stochastic dynamic programming framework that can capture the finite-sample behavior of a sampling policy. To implement OCBA in a fully sequential manner, they combine it with a ``most starving" sequential rule. Our proposed tree policy removes the known and bounded assumption of the node value, and balances exploration and exploitation to efficiently identify the optimal action. We demonstrate the efficiency of our new tree policy through numerical experiments on Tic-tac-toe and Gomoku.

The rest of the paper is organised as follows. Section \ref{sec:prob} formulates the proposed problem. The new tree policy and convergence results are proposed in Section 3. Section 4 provides numerical results. The last section concludes the paper.

\section{PROBLEM FORMULATION}\label{sec:prob}

We consider the setup of a finite-horizon discrete-time Markov decision process (MDP). An MDP is described by a four-tuple $\left(\mathcal{S},\mathcal{A},\mathcal{P},\mathcal{R}\right)$ with a horizon length $H$, where $\mathcal{S}$ is the set of states, $\mathcal{A}$ is the set of actions, $\mathcal{P} \equiv \mathcal{P}\left( {\left. {s'} \right|s,a} \right)$ is the Markovian transition kernel, $\mathcal{R} \equiv \mathcal{R}\left(s,a\right)$ is a random bounded reward function. The random reward can be discrete (win/draw/loss), continuous or a vector of reward values relative to each agent for more complex multi-agent domains. We assume that $\mathcal{A}$ and $\mathcal{S}$ are finite sets and $\mathcal{P}$ is deterministic, i.e., $\mathcal{P} \equiv \mathcal{P}\left( {\left. {s'} \right|s,a} \right) \in \left\{0,1\right\}$, $\forall s,s' \in \mathcal{S}$, $a \in \mathcal{A}$. The assumption of deterministic transition is reasonable since traditional MCTS is introduced in the context of deterministic games with a tree representation. At each stage, the system is in state $s \in \mathcal{S}$. After taking an action $a \in \mathcal{A}$, the state transits to next state $s' \in \mathcal{S}$ and an immediate reward is generated according to $\mathcal{R}\left(s,a\right)$. A stationary policy $\pi \left( {\left. a \right|s} \right)$ specifies the probability of performing action $a \in \mathcal{A}$ given current state $s \in \mathcal{S}$. The value function for each state $s \in \mathcal{S}$ under policy $\pi$ is defined as ${V^\pi }\left( s \right) = {\mathbb{E}_\pi }\left[ {\left. {\sum\nolimits_{t = 0}^{H - 1}  {\mathcal{R}\left( {{s_t},{a_t}} \right)}} \right|{s_0} = s}  \right]$. The state-action value function is defined as ${Q^\pi }\left( {s,a} \right) = {\mathbb{E}_\pi }\left[ {\left. {\sum\nolimits_{t = 0}^{H - 1} {\mathcal{R}\left( {{s_t},{a_t}} \right)} } \right|{s_0} = s,{a_0} = a} \right]$. The optimal value function under the optimal policy ${\pi ^*}$ is defined as ${V^*}\left( s \right) = {V^{{\pi ^*}}}\left( s \right) = {\sup _\pi }{V^\pi }\left( s \right)$, $\forall s \in \mathcal{S}$. The following Bellman equation holds: ${V^*}\left( s \right) = {\max _{a \in \mathcal{A}}}\left[ {\mathbb{E}\left[ {\mathcal{R}\left( {s,a} \right)} \right] + {V^*}\left( s^{\prime} \right)} \right] = {\max _{a \in \mathcal{A}}}{Q}\left( {s,a} \right)$, where $s^{\prime}$ is the next state reached by applying action $a$ on state $s$.

For the tree search problem, let $\mathbf{s}_{i} \mathop  = \limits^\Delta \left(s,i\right) \in \mathcal{S}$ and $\mathbf{a}_{i} \mathop  = \limits^\Delta \left(a,i\right) \in \mathcal{A}$ be a state $s \in \mathcal{S}$ and an action $a \in \mathcal{A}$ at depth $i$, $0 \le i < H$, respectively. We model the best action identification for every explored state node in the tree policy of MCTS as separate R\&S problems. All actions of current state node $\mathbf{s}_{i}$ are treated as alternatives. The optimal value of state node $V_i^*\left(\mathbf{s}_{i}\right)$ is unknown. Each state-action pair has an unknown value $Q_{i}\left(\mathbf{s}_i,\mathbf{a}_i\right)$, $0 \le i < H-1$, which is estimated by random samples $( {{\widehat Q}_{i}^1}\left(\mathbf{s}_i,\mathbf{a}_i\right), {{\widehat Q}_{i}^2}\left(\mathbf{s}_i,\mathbf{a}_i\right), \cdots ,{{\widehat Q}_{i}^{N_{i+1}\left(\mathbf{s}_{i+1}\right)}\left(\mathbf{s}_i,\mathbf{a}_i\right)} )$, where $N_{i+1}\left(\mathbf{s}_{i+1}\right) = \sum\nolimits_{\ell = 1}^t {\mathds{1}\left( {\mathbf{s}_{i+1} \in {\mathcal{P}^\ell}} \right)} \le T$ is the number of visits to the next state $\mathbf{s}_{i+1}$ after taking action $\mathbf{a}_{i}$ at state $\mathbf{s}_{i}$ in $1 \le t \le T$ roll-outs, $\mathcal{P}^\ell$ is the search path collected at the $\ell$-th roll-out, $T$ is total roll-outs or simulations (also known as the number of total simulation budget in the R\&S literature), and $\mathds{1}\left(\cdot\right)$ is an indicator function that equals to 1 when the event in the bracket is true and equals to 0 otherwise. We assume that ${{\widehat Q}_{i}^{\ell}}$, $0 \le i < H-1$, $1 \le \ell \le T$ are independent and identically distributed normal random variables, i.e., ${\widehat Q}_{i}^{\ell} \sim \mathcal{N} ( {Q_{i},{\sigma_{i} ^2}} )$ with a known state-action variance ${\sigma_i ^2}$, where we suppress $\left(\mathbf{s}_i,\mathbf{a}_i\right)$ for simplicity of notation. The sample variance is used as a plug-in for ${\sigma_i^2}$ in practice, i.e., ${\widehat \sigma }_i^2 = \frac{1}{{N_{i+1}( {\mathbf{s}_{i+1}} ) - 1}}\sum\nolimits_{\ell = 1}^{N_{i+1} ( {\mathbf{s}_{i+1}} )} {{( {{{\widehat Q}_{i}^{\ell}} - \bar Q_{i}} )^2}}$, where $\bar Q_{i} = \frac{1}{{N_{i+1} ( {\mathbf{s}_{i+1}})}}\sum\nolimits_{\ell = 1}^{N_{i+1} ( {\mathbf{s}_{i+1}} )} {{{\widehat Q}_{i}^{\ell}}}$ is sample mean. Under a Bayesian framework, we assume the prior distribution of $Q_{i}$ is a conjugate prior of the sampling distribution of $\widehat Q_i^{\ell}$, which is also a normal distribution $\mathcal{N} ( {{Q_i^{\left( 0 \right)}},( {{\sigma_i^{\left( 0 \right)}}} )^2} )$. Then the posterior distribution of $Q_i$ is $\mathcal{N} ( {{Q_i^{\left( t \right)}},( {{\sigma_i^{\left( t \right)}}} )^2} )$, $0 \le i < H-1$, $1 \le t \le T$, with posterior state-action variance
\begin{equation}\label{pv}
{\left( {\sigma_i^{\left( t \right)}}\left(\mathbf{s}_{i},\mathbf{a}_{i}\right) \right)^2} = {\left( {\frac{1}{{{{\left( {{\sigma_i^{\left( 0 \right)}}\left(\mathbf{s}_{i},\mathbf{a}_{i}\right)} \right)}^2}}} + \frac{{N_{i+1}\left( {\mathbf{s}_{i+1}} \right)}}{{{\sigma_i^2}\left(\mathbf{s}_{i},\mathbf{a}_{i}\right)}}} \right)^{ - 1}}~,
\end{equation}
and posterior state-action mean
\begin{equation}\label{pm}
{Q_i^{\left( t \right)}}\left(\mathbf{s}_{i},\mathbf{a}_{i}\right) = {\left( {{\sigma_i^{\left( t \right)}}\left(\mathbf{s}_{i},\mathbf{a}_{i}\right)} \right)^2}{\left( {\frac{{{Q_i^{\left( 0 \right)}}\left(\mathbf{s}_{i},\mathbf{a}_{i}\right)}}{{{{\left( {{\sigma_i^{\left( 0 \right)}}\left(\mathbf{s}_{i},\mathbf{a}_{i}\right)} \right)}^2}}} + \frac{{N_{i+1}\left( {\mathbf{s}_{i+1}} \right){\bar Q_i}\left(\mathbf{s}_{i},\mathbf{a}_{i}\right)}}{{{\sigma_i^2}\left(\mathbf{s}_{i},\mathbf{a}_{i}\right)}}} \right)^{ - 1}}~.
\end{equation}

Note that if $\sigma_i^{\left( 0 \right)}\left(\mathbf{s}_{i},\mathbf{a}_{i}\right) \to \infty$, then $Q_i^{\left( t \right)}\left(\mathbf{s}_{i},\mathbf{a}_{i}\right) = \bar Q_i\left(\mathbf{s}_{i},\mathbf{a}_{i}\right)$ and $( {{\sigma_i^{\left( t \right)}}\left(\mathbf{s}_{i},\mathbf{a}_{i}\right)} )^2 = {{\sigma_i^2\left(\mathbf{s}_{i},\mathbf{a}_{i}\right)} / {{N_{i+1}\left( {\mathbf{s}_{i+1}} \right)}}}$ and such a case is called uninformative. We aim to identify the best action that achieves the highest state-action value at the initial state $\mathbf{s}_0$, i.e., finding $\mathbf{a}_0^* = \arg \mathop {\max }\nolimits_{{\mathbf{a}_0} \in {\mathcal{A}_{\mathbf{s}_0}}} Q_0\left( {{\mathbf{s}_0},{\mathbf{a}_0}} \right)$, where $\mathcal{A}_{\mathbf{s}_0}$ is the set of actions at state $\mathbf{s}_0$. A correct selection of the best action occurs when $(\mathbf{a}_0^{\left( T \right)} )^{*} = \mathbf{a}_0^*$, where $(\mathbf{a}_0^{\left( T \right)} )^{*} = \arg {\max _{{\mathbf{a}_0} \in \mathcal{A}_{\mathbf{s}_0}}}{Q_0^{\left( T \right)}}\left( {{\mathbf{s}_0},{\mathbf{a}_0}} \right)$ is the estimated best action that achieves the highest posterior mean at the initial state $\mathbf{s}_0$ after $T$ roll-outs. The PCS for selecting $\mathbf{a}_0^*$ can be expressed as
 $${\rm PCS} = \Pr \left\{ {\bigcap\nolimits_{a \in \mathcal{A}_{\mathbf{s}_0},\; a \ne (\mathbf{a}_0^{\left( T \right)} )^{*}} {{Q_0^{\left( T \right)}} ( {{\mathbf{s}_0},(\mathbf{a}_0^{\left( T \right)} )^{*}} ) > {Q_0^{\left( T \right)}}\left( {{\mathbf{s}_0},a} \right)} } \right\}~.$$

 We aim to find an efficient dynamic sampling tree policy such that the ${\rm PCS}$ can be maximized. Compared with minimizing the expected cumulative regret in the canonical MAB problem, maximizing ${\rm PCS}$ results in an allocation of limited computational budget in a way that optimally balances exploration and exploitation. Based on the information collected from simulations, sampling policy ${\mathbf{A}_T}\left(  \cdot  \right) \mathop  = \limits^\Delta \left( {{\mathds{A}_1}\left(  \cdot  \right),{\mathds{A}_2}\left(  \cdot  \right), \cdots ,{\mathds{A}_T}\left(  \cdot  \right)} \right)$ is a sequence of mappings, where ${\mathds{A}_t}\left( \cdot \right) \in \{{\mathbf{a}_0^1},{\mathbf{a}_0^2},\cdots,{\mathbf{a}_0^{ | {{\mathcal{A}_{{\mathbf{s}_0}}}} |}}\}$ allocates the $t$-th computational budget to an action of the initial state based on the information $\mathcal{E} _{t - 1}$ collected throughout the first $\left( {t - 1} \right)$ roll-outs, where $\left|  \cdot  \right|$ denotes the cardinality of a set. The expected payoff for a dynamic sampling tree policy can be recursively defined in a stochastic dynamic programming problem by
$${\mathcal{V}_T}\left( {\mathcal{E} _T};{\mathbf{A}_T} \right) \mathop  = \limits^\Delta \Pr \left\{ {\left. {\bigcap\nolimits_{a \in \mathcal{A}_{\mathbf{s}_0},\; a \ne (\mathbf{a}_0^{\left( T \right)} )^{*}} {{Q_0^{\left( T \right)}} ( {{\mathbf{s}_0},(\mathbf{a}_0^{\left( T \right)} )^{*}} ) > {Q_0^{\left( T \right)}}\left( {{\mathbf{s}_0},a} \right)} } \right|{\mathcal{E} _T}} \right\}~,$$
and for $0 \le t < T$,
$${\mathcal{V}_t}\left( {{{\mathcal{E}} _t};\mathbf{A}_T} \right) \mathop  = \limits^\Delta \mathbb{E}{\left. {\left[ {\left. {{\mathcal{V}_{t + 1}}\left( {{\mathcal{E} _t} \cup \left\{ {\widehat Q}_0^{\left({N_1\left(\mathbf{s}_1^i\right) +1}\right)}\left(\mathbf{s}_0,\mathbf{a}_0^i\right) \right\};{\mathcal{A}_T\left(\cdot\right)}} \right)} \right|{\mathcal{E} _t}} \right]} \right|_{{\mathbf{a}_0^i} = {\mathds{A}_{t + 1}}\left( {{\mathcal{E} _t}} \right)}}~,$$
where ${\widehat Q}_0^{\left({N_1\left(\mathbf{s}_1^i\right) +1}\right)}\left(\mathbf{s}_0,\mathbf{a}_0^i\right)$ is the $\left({N_1\left(\mathbf{s}_1^i\right) +1}\right)$-th sample for allocated action $\mathbf{a}_0^i \in \mathcal{A}_{\mathbf{s}_0}$. Then an optimal dynamic sampling tree policy can be defined as the solution of the stochastic dynamic programming problem: $\mathbf{A}_T^*\mathop  = \limits^\Delta \arg \mathop {\max }\nolimits_{\mathbf{A}_T} {\mathcal{V}_0}\left( {\mathcal{E}_0;\mathbf{A}_T} \right)$, where $\mathcal{E}_0$ contains the prior information. Such a stochastic dynamic programming problem can be viewed as a MDP, and then the optimality condition of a dynamic sampling tree policy is governed by the Bellman equation of the MDP. However, solving such a MDP typically suffers from curse-of-dimensionality. In the R\&S literature, \cite{peng2018ranking} find a suitable value function approximation (VFA) for the Bellman equations and use a further approximation for the VFA, which leads to the so-called AOAP algorithm that maximizes a VFA one-step look ahead. Inspired by their work, we propose a tree policy by applying the AOAP algorithm to each node of the tree, leading to a dynamic sampling tree policy for MCTS.

\section{A NEW TREE POLICY}

In this section, we first briefly describe the AOAP algorithm under the tree search setup. Then we propose a new tree policy for MCTS that finds the best action at each state node.

In the tree policy of MCTS, for each visited state node $\mathbf{s}_i$ in the search path at the $t$-th roll-out, the AOAP algorithm first identifies the action with the largest posterior state-action mean $(\mathbf{a}_{i}^{\left( t \right)})^{*} = \arg {\max _{\mathbf{a}_i \in \mathcal{A}_{\mathbf{s}_i}}}{Q_i^{\left( t \right)}}\left( {\mathbf{s}_i,\mathbf{a}_i} \right)$, and then calculates the following equations:
\begin{equation}\label{equ21}
{\mathcal{\widetilde V}_t}\left( \mathbf{s}_i, (\mathbf{a}_{i}^{\left( t \right)})^{*} \right) = \mathop {\min }\limits_{\mathbf{a}_i \ne (\mathbf{a}_{i}^{\left( t \right)})^{*}} \frac{{{{\left( {Q_i^{\left( t \right)}\left( {\mathbf{s}_i,(\mathbf{a}_{i}^{\left( t \right)})^{*}} \right) - Q_i^{\left( t \right)}\left( {\mathbf{s}_i,\mathbf{a}_i} \right)} \right)}^2}}}{{{{\left( {\sigma_i^{\left( {t + 1} \right)}\left( {\mathbf{s}_i,(\mathbf{a}_{i}^{\left( t \right)})^{*}} \right)} \right)}^2} + {{\left( {\sigma_i^{\left( t \right)}\left( {\mathbf{s}_i,\mathbf{a}_i} \right)} \right)}^2}}}~,
\end{equation}
and for ${\mathbf{a}_i,{\mathbf{\widetilde  a}}_i \ne (\mathbf{a}_{i}^{\left( t \right)})^{*}}$, $\mathbf{a}_i,{\mathbf{\widetilde  a}}_i \in \mathcal{A}_{\mathbf{s}_i}$,
\begin{equation}\label{equ22}
\begin{aligned}
{\mathcal{\widetilde V}_t}\left( \mathbf{s}_i,\mathbf{a}_i \right) = \min  & \left\{ \frac{{{{\left( {Q_i^{\left( t \right)}\left( {\mathbf{s}_i,(\mathbf{a}_{i}^{\left( t \right)})^{*}} \right) - Q_i^{\left( t \right)}\left( {\mathbf{s}_i,\mathbf{a}_i} \right)} \right)}^2}}}{{{{\left( {\sigma_i^{\left( t \right)}\left( {\mathbf{s}_i,(\mathbf{a}_{i}^{\left( t \right)})^{*}} \right)} \right)}^2} + {{\left( {\sigma_i^{\left( {t + 1} \right)}\left( {\mathbf{s}_i,\mathbf{a}_i} \right)} \right)}^2}}}, \right.\\
& \left.\mathop {\min }\limits_{{\mathbf{\widetilde  a}}_i \ne \mathbf{a}_i,(\mathbf{a}_{i}^{\left( t \right)})^{*}} \frac{{{{\left( {Q_i^{\left( t \right)}\left( {\mathbf{s}_i,(\mathbf{a}_{i}^{\left( t \right)})^{*}} \right) - Q_i^{\left( t \right)}\left( {\mathbf{s}_i,{\mathbf{\widetilde  a}}_i} \right)} \right)}^2}}}{{{{\left( {\sigma_i^{\left( t \right)}\left( {\mathbf{s}_i,(\mathbf{a}_{i}^{\left( t \right)})^{*}} \right)} \right)}^2} + {{\left( {\sigma_i^{\left( t \right)}\left( {\mathbf{s}_i,{\mathbf{\widetilde  a}}_i} \right)} \right)}^2}}} \right\}~,
\end{aligned}
\end{equation}
where
$${\left( {{\sigma_i^{\left( {t + 1} \right)}}\left( {\mathbf{s}_i,\mathbf{a}_i} \right)} \right)^2} = {\left( {\frac{1}{{{{\left( {{\sigma_i^{\left( 0 \right)}}\left( {\mathbf{s}_i,\mathbf{a}_i} \right)} \right)}^2}}} + \frac{{N_{i+1}\left( {\mathbf{s}_{i+1}} \right) + 1}}{{{\sigma_i^2}\left( {\mathbf{s}_i,\mathbf{a}_i} \right)}}} \right)^{ - 1}}~.$$

After calculating values of ${\mathcal{\widetilde V}_t}\left( \mathbf{s}_i,\mathbf{a}_i\right)$, $\forall~\mathbf{a}_i \in \mathcal{A}_{\mathbf{s}_i}$, the AOAP algorithm selects the action with the largest ${\mathcal{\widetilde V}_t}\left( \mathbf{s}_i,\mathbf{a}_i\right)$, i.e., sample
\begin{equation}\label{AOAPalg}
    \mathbf{\widehat a}_{i}^{\left(t\right)} = \arg \mathop {\max }\nolimits_{\mathbf{a}_i \in \mathcal{A}_{\mathbf{s}_i}}{\mathcal{\widetilde V}_t}\left( \mathbf{s}_i,\mathbf{a}_i\right)~.
\end{equation}

The MCTS algorithm using the AOAP as a tree policy is named as AOAP-MCTS. Compared with UCT, the tree policy based on AOAP utilizes posterior means and posterior variances, which incorporate average value, variances and the number of visits of nodes. The proposed tree policy attempts to balance the exploration of nodes with high variances and exploitation of nodes with high state-action values. In implementation, if more than one action has the same maximal posterior state-action mean or has the same value of ${\mathcal{\widetilde V}_t}\left( \mathbf{s}_i,\mathbf{a}_i\right)$, the tie can be broken by choosing randomly or the action with the highest ${{{( {{\sigma_i ^{\left( t \right)}}\left( {\mathbf{s}_i,\mathbf{a}_i} \right)} )^2}} / {N_{i+1}\left( {\mathbf{s}_{i+1}} \right)}}$, that is, choosing the action with low frequency of visits and large posterior state-action variance. In addition, notice that we use variance information of a node as a denominator in calculation of both posterior state-action mean and variance, and in order to ensure the variance is positive, a small positive real number $\epsilon$ can be introduced when ${{\widehat \sigma}_i^2}\left( {\mathbf{s}_i,\mathbf{a}_i} \right) = 0$.

We highlight some major modifications to the canonical MCTS when using AOAP as the tree policy. First, ${\bar Q_i}\left( {\mathbf{s}_i,\mathbf{a}_i} \right)$, ${{\widehat \sigma_i} ^2}\left( {\mathbf{s}_i,\mathbf{a}_i} \right)$ and $N_{i+1}\left( {\mathbf{s}_{i+1}} \right)$, $0 \le i < H-1$ are required to store for each node in the tree. The prior information $Q_i^{\left( 0 \right)}\left(\mathbf{s}_{i},\mathbf{a}_{i}\right)$ and $\sigma_i^{\left( 0 \right)}\left(\mathbf{s}_{i},\mathbf{a}_{i}\right)$ can be specified and adjusted in implementation. Second, in order to calculate ${{\widehat \sigma_i}^2}\left( {\mathbf{s}_i,\mathbf{a}_{i}} \right)$ for each state-action node, each state node $\mathbf{s}_i$ is required to be well-expanded when it is visited, and each state-action pair $\left( {\mathbf{s}_i,\mathbf{a}_{i}} \right)$, $\forall~\mathbf{a}_{i} \in \mathcal{A}_{\mathbf{s}_i}$ and its corresponding child state-action node $\mathbf{s}_{i+1}$ is required to be added to the search path. Each state-action pair is required to be sampled $n_0 > 1$ times, i.e., a state node is expandable when one of its child nodes is visited less than $n_0$ times. Third, after receiving the terminal reward $\Delta_{t}$ of the collected search path $\mathcal{P}^{t}$ at $t$-th roll-out, all values of nodes in the collected search path are updated in reversed order through: for $0 \le i < \ell$ and let $\mathbf{s}_{\ell}$ be the leaf node in $\mathcal{P}^{t}$,
\begin{equation}\label{up1}
    {N_{i+1}}\left( {\mathbf{s}_{i+1}} \right) \leftarrow {N_{i+1}}\left( {\mathbf{s}_{i+1}} \right) + 1~,
\end{equation}
\begin{equation}\label{up2}
    \widehat V_\ell^*\left( {{\mathbf{s}_\ell}} \right) \leftarrow \frac{{{N_\ell}\left( {{\mathbf{s}_\ell}} \right) - 1}}{{{N_\ell}\left( {{\mathbf{s}_\ell}} \right)}}\hat V_\ell^*\left( {{\mathbf{s}_\ell}} \right) + \frac{1}{{{N_\ell}\left( {{\mathbf{s}_\ell}} \right)}}{\Delta _t}~,
\end{equation}
\begin{equation}\label{up3}
  \widehat Q_i^{{N_{i + 1}}\left( {{\mathbf{s}_{i + 1}}} \right)}\left( {{\mathbf{s}_i},{\mathbf{a}_i}} \right) = R\left( {{\mathbf{s}_i},{\mathbf{a}_i}} \right) + \hat V_{i + 1}^*\left( {{\mathbf{s}_{i + 1}}} \right)~,
\end{equation}
\begin{equation}\label{up4}
    {{\bar \mu}_i}\left( {\mathbf{s}_{i},\mathbf{a}_i} \right) = {{\bar Q}_i}\left( {\mathbf{s}_{i},\mathbf{a}_i} \right)~,
\end{equation}
\begin{equation}\label{up5}
    {{\bar Q}_{i}}\left( {\mathbf{s}_{i},\mathbf{a}_i} \right) \leftarrow \frac{{{N_{i+1}}\left( {\mathbf{s}_{i+1}} \right) - 1}}{{{N_{i+1}}\left( {\mathbf{s}_{i+1}} \right)}}{{\bar Q}_i}\left( {\mathbf{s}_{i},\mathbf{a}_i} \right) + \frac{1}{{{N_{i+1}}\left( {\mathbf{s}_{i+1}} \right)}}{\Delta _t}~,
\end{equation}
\begin{equation}\label{up6}
    \widehat \sigma _{i}^2\left( {\mathbf{s}_i,\mathbf{a}_i} \right) \leftarrow \frac{{{N_{i+1}}\left( {\mathbf{s}_{i+1}} \right) - 1}}{{{N_{i + 1}}\left( {\mathbf{s}_{i+1}} \right)}}\widehat \sigma _i^2\left( {\mathbf{s}_i,\mathbf{a}_i} \right) + \frac{1}{{{N_{i + 1}}\left( {\mathbf{s}_{i+1}} \right)}}\left( {{\Delta _t} - {{\bar Q}_i}\left( {\mathbf{s}_i,\mathbf{a}_i} \right)} \right)\left( {{\Delta _t} - {{\bar \mu}_{i}}\left( {\mathbf{s}_i,\mathbf{a}_i} \right)} \right)~,
\end{equation}
\begin{equation}\label{up7}
    \widehat V_i^*\left( {{\mathbf{s}_i}} \right) \leftarrow \mathop {\max }\nolimits_{{\mathbf{a}_i} \in {\mathcal{A}_{{\mathbf{s}_i}}}}\bar Q\left( {{\mathbf{s}_i},{\mathbf{a}_i}} \right)~.
\end{equation}

Algorithm \ref{AOAP-MCTSpseudo} shows the pseudocode of the AOAP-MCTS algorithm. The AOAP-MCTS algorithm is run with $T$ roll-outs from the root state node $\mathbf{s}_0$, after which a game tree is built and the estimated optimal action $(\mathbf{a}_0^{\left( T \right)} )^{*}$ is found corresponding to an action of the root node with the highest posterior state-action mean. Notice that since we consider deterministic transitions, the tree is fixed once the root node is chosen. When a node in the tree is visited, the tree policy first determines whether the node is expandable. If there are state-action pairs that are not yet part of the tree, one of those is chosen randomly and added to the tree, and if there are state-action pairs are visited less than $n_0$ times, one of those is chosen randomly. If all state-action pairs are well-expanded, AOAP is used to find the allocated one. $\mathbf{s}_{\ell}$ is the node reached during the tree policy stage corresponding to state $s$ at depth $\ell$. A simulation is run from $\mathbf{s}_{\ell}$ according to a default policy, until a terminal node has been reached. The reward $\Delta_t$ of the terminal state is then backpropagated to all nodes collected in the search path during this round to update the node statistics.

\begin{breakablealgorithm}\label{AOAP-MCTSpseudo}
 \caption{The AOAP-MCTS algoritm}
 \begin{algorithmic}
 \Require root state node $\mathbf{s}_0$, number of roll-outs $T$, and algorithmic constants: $n_0$, $\epsilon$, $Q_i^{\left( 0 \right)}\left(\mathbf{s}_{i},\mathbf{a}_{i}\right)$ and $\sigma_i^{\left( 0 \right)}\left(\mathbf{s}_{i},\mathbf{a}_{i}\right)$.
\Ensure $(\mathbf{a}_0^{\left( T \right)} )^{*}$ \\
\Function {AOAP-MCTS}{$\mathbf{s}_0$, $T$, $n_0$, $\epsilon$, $Q_i^{\left( 0 \right)}\left(\mathbf{s}_{i},\mathbf{a}_{i}\right)$, $\sigma_i^{\left( 0 \right)}\left(\mathbf{s}_{i},\mathbf{a}_{i}\right)$}
\State $t \gets 0$
\While{$t < T$}
\State $\mathbf{s}_{\ell} \gets$ \Call{TREEPOLICY}{$\mathbf{s}_{0}$, $n_0$, $\epsilon$, $Q_i^{\left( 0 \right)}\left(\mathbf{s}_{i},\mathbf{a}_{i}\right)$, $\sigma_i^{\left( 0 \right)}\left(\mathbf{s}_{i},\mathbf{a}_{i}\right)$}
\State $\Delta_t \gets$ \Call{DEFAULTPOLICY}{$\mathbf{s}_{\ell}$}
\State \Call{BACKPROPAGATE}{$\mathbf{s}_{\ell}$, $\Delta_t$}
\State $t \gets {t+1}$
\EndWhile
\State \Return{$(\mathbf{a}_0^{\left( T \right)} )^{*}$}
\EndFunction
\State
 \Function{TREEPOLICY}{$\mathbf{s}_i$}
  \While{$\mathbf{s}_i$ is nonterminal}
   \If{$\mathbf{s}_i$ is expandable}
    \State \Return{\Call{EXPAND}{$\mathbf{s}_i$}}
   \Else
    \State $\mathbf{s}_{i+1} \gets$ \Call{AOAP}{$\mathbf{s}_i$, $n_0$, $\epsilon$, $Q_i^{\left( 0 \right)}\left(\mathbf{s}_{i},\mathbf{a}_{i}\right)$, $\sigma_i^{\left( 0 \right)}\left(\mathbf{s}_{i},\mathbf{a}_{i}\right)$}
   \EndIf
 \EndWhile
 \State \Return{$\mathbf{s}_{i+1}$}
 \EndFunction
 \State
 \Function{EXPAND}{$\mathbf{s}_i$} \\
  \quad choose $\mathbf{a}_i \in$ untried actions from $\mathcal{A}_{\mathbf{s}_i}$ \\
  \quad append a new child $\mathbf{s}_{i+1}$ to $\mathbf{s}_i$
 \State \Return{$\mathbf{s}_{i+1}$}
 \EndFunction
 \State
 \Function{AOAP}{$\mathbf{s}_i$, $n_0$, $\epsilon$, $Q_i^{\left( 0 \right)}\left(\mathbf{s}_{i},\mathbf{a}_{i}\right)$, $\sigma_i^{\left( 0 \right)}\left(\mathbf{s}_{i},\mathbf{a}_{i}\right)$} \\
 \quad by solving equations (\ref{pv}), (\ref{pm}), (\ref{equ21}), (\ref{equ22}) and (\ref{AOAPalg})
 \State \Return allocated action $\mathbf{\widehat a}_{i}^{\left(t\right)}$
 \EndFunction
  \State
 \Function{DEFAULTPOLICY}{$\mathbf{s}_{\ell}$}
 \While{$\mathbf{s}_j$, $ j \ge \ell+1$ is nonterminal} \\
 \quad \quad \quad choose $\mathbf{a}_j \in \mathcal{A}\left(\mathbf{s}_j\right)$ uniformly at random
 \State append a new child $\mathbf{s}_{j+1}$ to $\mathbf{s}_j$
 \EndWhile
 \State \Return{reward $\Delta_t$ of the collected search path}
 \EndFunction
 \State
 \Function{BACKPROPAGATE}{$\mathbf{s}_{\ell}$,$\Delta_t$}
  \While{$\mathbf{s}_i$, $0 \le i \le \ell$ is not null} \\
  \quad \quad \quad update node values through equations (\ref{up1}), (\ref{up2}), (\ref{up3}), (\ref{up4}), (\ref{up5}), (\ref{up6}) and (\ref{up7})
  \EndWhile
  \EndFunction
 \end{algorithmic}
\end{breakablealgorithm}

We show theoretical results regarding AOAP-MCTS.
\begin{prop}
The proposed AOAP-MCTS is consistent, i.e., $\mathop {\lim }\limits_{T \to \infty }{ ( {\mathbf{a}_i^{\left( T \right)}} )^*} = a_i^*$, $\mathop {\lim }\limits_{T \to \infty } \widehat V_i^*\left( {{\mathbf{s}_i}} \right) = V_i^*\left( {{\mathbf{s}_i}} \right)$, $0 \le i < H$.
\end{prop}

At each explored state node in the tree policy, the best action is identified by the AOAP algorithm. As shown in \cite{peng2018ranking}, AOAP is consistent, i.e., every alternative will be sampled infinitely often almost surely as the number of computational budget goes to infinity, so that the best alternative can be definitely selected. Following their analysis, Proposition 1 can be proved by induction. We leave the proof to future work.

\section{NUMERICAL EXPERIMENTS}
In this section, we conduct numerical experiments to test the performances of different tree policies for MCTS. We apply our proposed algorithm to the games of Tic-tac-toe and Gomoku. The proposed AOAP-MCTS is compared with UCT in \cite{kocsis2006bandit}, OCBA-MCTS in \cite{li2021optimal} and TTTS-MCTS, which runs tree policy by the Top-Two Thompson Sampling (TTTS) in \cite{russo2020simple}. We describe the three tree policies as follows:
\begin{itemize}
\item UCT: The policy selects the action with the highest upper confidence bound, i.e.,
$$\widehat {\mathbf{a}}_i^{\left( t \right)} = \arg {\max\nolimits_{{\mathbf{a}_i} \in {\mathcal{A}_{{\mathbf{s}_i}}}}}\left\{{{\bar Q}_i}\left( {{\mathbf{s}_i},{\mathbf{a}_i}} \right) + {C_p}\sqrt {{{2\log {N_i}\left( {{\mathbf{s}_i}} \right)}} / {{{N_{i + 1}}\left( {{\mathbf{s}_{i + 1}}} \right)}}}\right\}~,$$
where $C_p$ is the exploration weight. We choose $C_p = 1$ in implementation.
\item OCBA-MCTS: The policy solves a set of equations and selects the action that is the most starving, i.e., $\forall {\mathbf{a}_i},{{ \mathbf{\widetilde a}}_i} \ne  \mathbf{\widehat a}_i^*,{\mathbf{a}_i},{{ \mathbf{\widetilde a}}_i} \in {\mathcal{A}_{{\mathbf{s}_i}}}$, let $\mathbf{\widehat a}_i^* = \arg {\max _{{\mathbf{a}_i} \in {\mathcal{A}_{{\mathbf{s}_i}}}}}{{\bar Q}_i}\left( {{\mathbf{s}_i},{\mathbf{a}_i}} \right)$ and ${\delta _i}\left( {\mathbf{\widehat  a}_i^*,{\mathbf{a}_i}} \right) = {{\bar Q}_i}\left( {{\mathbf{s}_i}, \mathbf{\widehat a}_i^*} \right) - {{\bar Q}_i}\left( {{\mathbf{s}_i},{\mathbf{a}_i}} \right)$, $\forall {\mathbf{a}_i} \ne \mathbf{\widehat a}_i^*$,
$$\frac{{{{\widetilde N}_{i + 1}}\left( {{\mathbf{s}_{i + 1}}} \right)}}{{{{\widetilde N}_{i + 1}}\left( {{{ \mathbf{\widetilde s}}_{i + 1}}} \right)}} = {\left( {\frac{{{{{\sigma _i}\left( {{\mathbf{s}_i},{\mathbf{a}_i}} \right)} / {{\delta _i}\left( { \mathbf{\widehat a}_i^*,{\mathbf{a}_i}} \right)}}}}{{{{{\sigma _i}\left( {{\mathbf{s}_i},{{\mathbf{\widetilde a}}_i}} \right)} / {{\delta _i}\left( { \mathbf{\widehat a}_i^*,{{ \mathbf{\widetilde a}}_i}} \right)}}}}} \right)^2}~,$$
$${{\widetilde N}_{i + 1}}\left( {\mathbf{\mathbf{\widetilde  s}}_{i + 1}^*} \right) = {\sigma _i}\left( {{\mathbf{s}_i}, \mathbf{\widehat a}_i^*} \right)\sqrt {\sum\nolimits_{{\mathbf{a}_i} \in {\mathcal{A}_{{\mathbf{s}_i}}},{\mathbf{a}_i} \ne  \mathbf{\widehat a}_i^*} {\frac{{{{\left( {{{\widetilde N}_{i + 1}}\left( {{\mathbf{s}_{i + 1}}} \right)} \right)}^2}}}{{\sigma _i^2\left( {{\mathbf{s}_i},{\mathbf{a}_i}} \right)}}} }~,$$
$$ \mathbf{\widehat a}_i^{\left( t \right)} = \arg {\max\nolimits _{{\mathbf{a}_i} \in {\mathcal{A}_{{\mathbf{s}_i}}}}}\left( {{{\widetilde N}_{i + 1}}\left( {{\mathbf{s}_{i + 1}}} \right) - {N_{i + 1}}\left( {{\mathbf{s}_{i + 1}}} \right)} \right)~.$$
\item TTTS-MCTS: The policy first samples $\widehat Q_i^1\left( {{\mathbf{s}_i},{\mathbf{a}_i}} \right)$, $\forall \mathbf{a}_i \in \mathcal{A}_{\mathbf{s}_i}$ from $\mathcal{N}( {{Q_i^{\left( t \right)}\left(\mathbf{s}_i,\mathbf{a}_i\right)},( {{\sigma_i^{\left( t \right)}\left(\mathbf{s}_i,\mathbf{a}_i\right)}} )^2} )$, and finds $ \mathbf{\widehat a}_{i1}^{\left( t \right)} = \arg {\max _{{\mathbf{a}_i} \in {\mathcal{A}_{{\mathbf{s}_i}}}}}\widehat Q_i^1\left( {{\mathbf{s}_i},{\mathbf{a}_i}} \right)$. Then the policy samples $\widehat Q_i^2\left( {{\mathbf{s}_i},{\mathbf{a}_i}} \right)$, $\forall \mathbf{a}_i \in \mathcal{A}_{\mathbf{s}_i}$ from the same distribution until $\mathbf{\widehat a}_{i1}^{\left( t \right)} \ne \mathbf{\widehat a}_{i2}^{\left( t \right)}$, where $ \mathbf{\widehat a}_{i2}^{\left( t \right)} = \arg {\max _{{\mathbf{a}_i} \in {\mathcal{A}_{{\mathbf{s}_i}}}}}\hat Q_i^2\left( {{\mathbf{s}_i},{\mathbf{a}_i}} \right)$. The allocated action is determined by randomly choosing from $\mathbf{\widehat a}_{i1}^{\left( t \right)}$ and $ \mathbf{\widehat a}_{i2}^{\left( t \right)}$. Since the second stage of the policy can be time-consuming when the action space is large, we truncate it with 10 rounds in implementation, i.e., if $ \mathbf{\widehat a}_{i2}^{\left( t \right)}$ can not be found in 10 rounds, we determine $\mathbf{\widehat a}_{i2}^{\left( t \right)}$ by the second largest value of $\widehat Q_i^1\left( {{\mathbf{s}_i},{\mathbf{a}_i}} \right)$.
\end{itemize}

\textit{Experiment 1: Tic-tac-toe} Tic-tac-toe is a game played on a three-by-three board by two players, who alternately place the marks `X' and `O' in one of the nine spaces in the board. The player who succeeds in placing three of their marks in a horizontal, vertical, or diagonal row is the winner. If both players act optimally, the game will always end in a draw.

\begin{figure}[!h]
\centering
\subfigure[Setup 1]{
\includegraphics[width=0.22\textwidth]{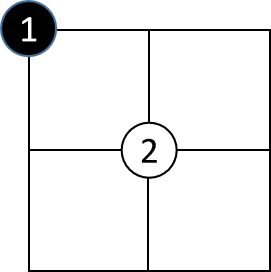}
}
\quad
\subfigure[Setup 2]{
\includegraphics[width=0.24\textwidth]{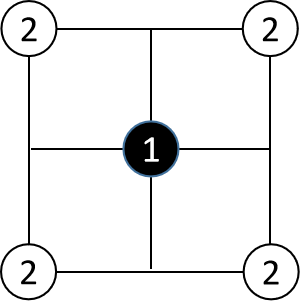}
}
\caption{Tic-tac-toe board setup in Experiment 1.1.}
\label{fig1.1boa}
\end{figure}

\textit{Experiment 1.1: Precision} In this experiment, we focus on the precision of MCTS in finding the optimal move under different tree policies. The effectiveness of a policy is measured by PCS. Given a place marked by Player 1, we apply different tree policies to identify the optimal move for Player 2. Figure \ref{fig1.1boa} shows two board setups, where we use black and white to represent `X' and `O', respectively, for ease of presentation.

\begin{figure}[!h]
\centering
\subfigure[Player 1 plays randomly in Setup 1]{
\includegraphics[width=0.43\textwidth]{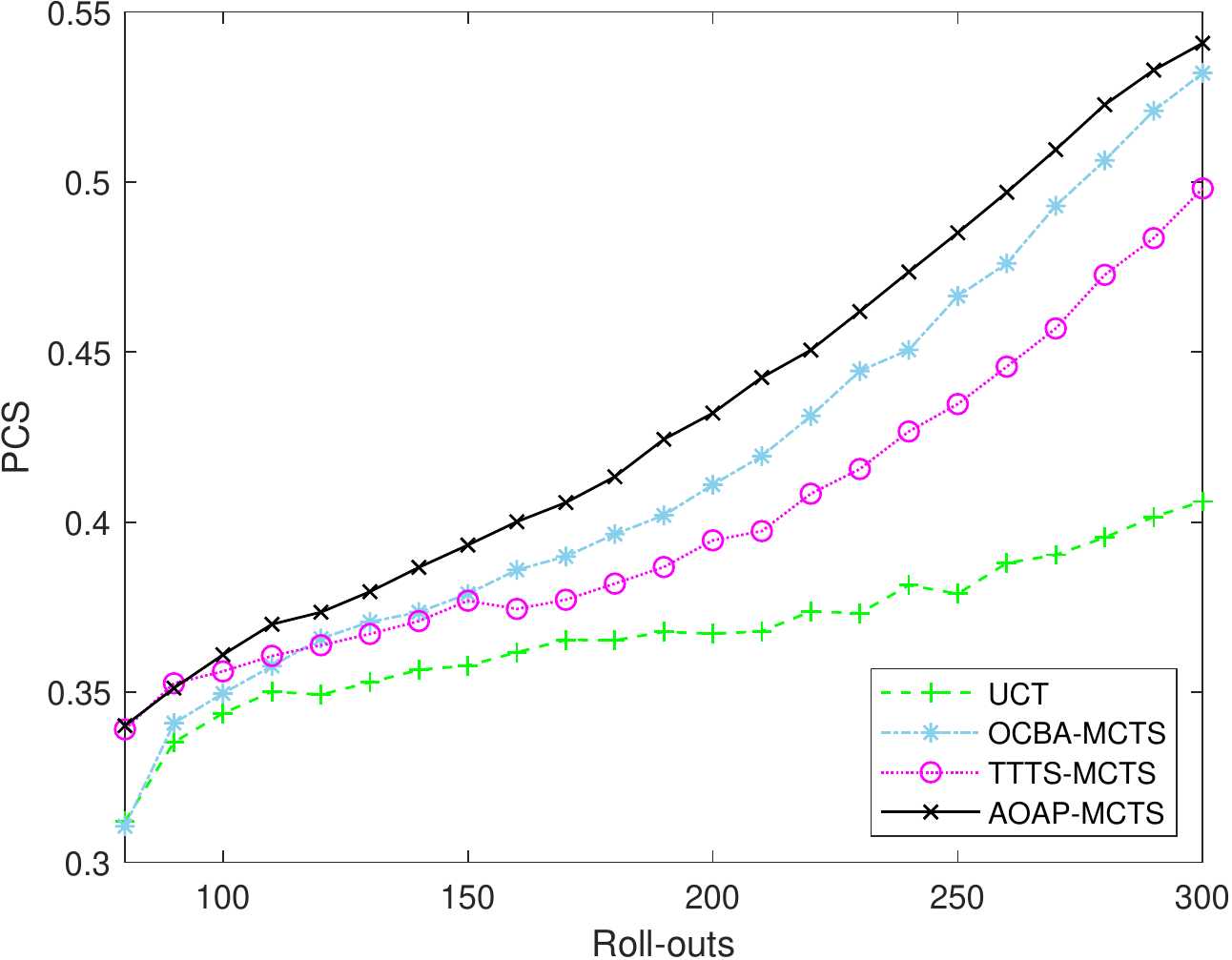}
}
\subfigure[Player 1 plays randomly in Setup 2]{
\includegraphics[width=0.43\textwidth]{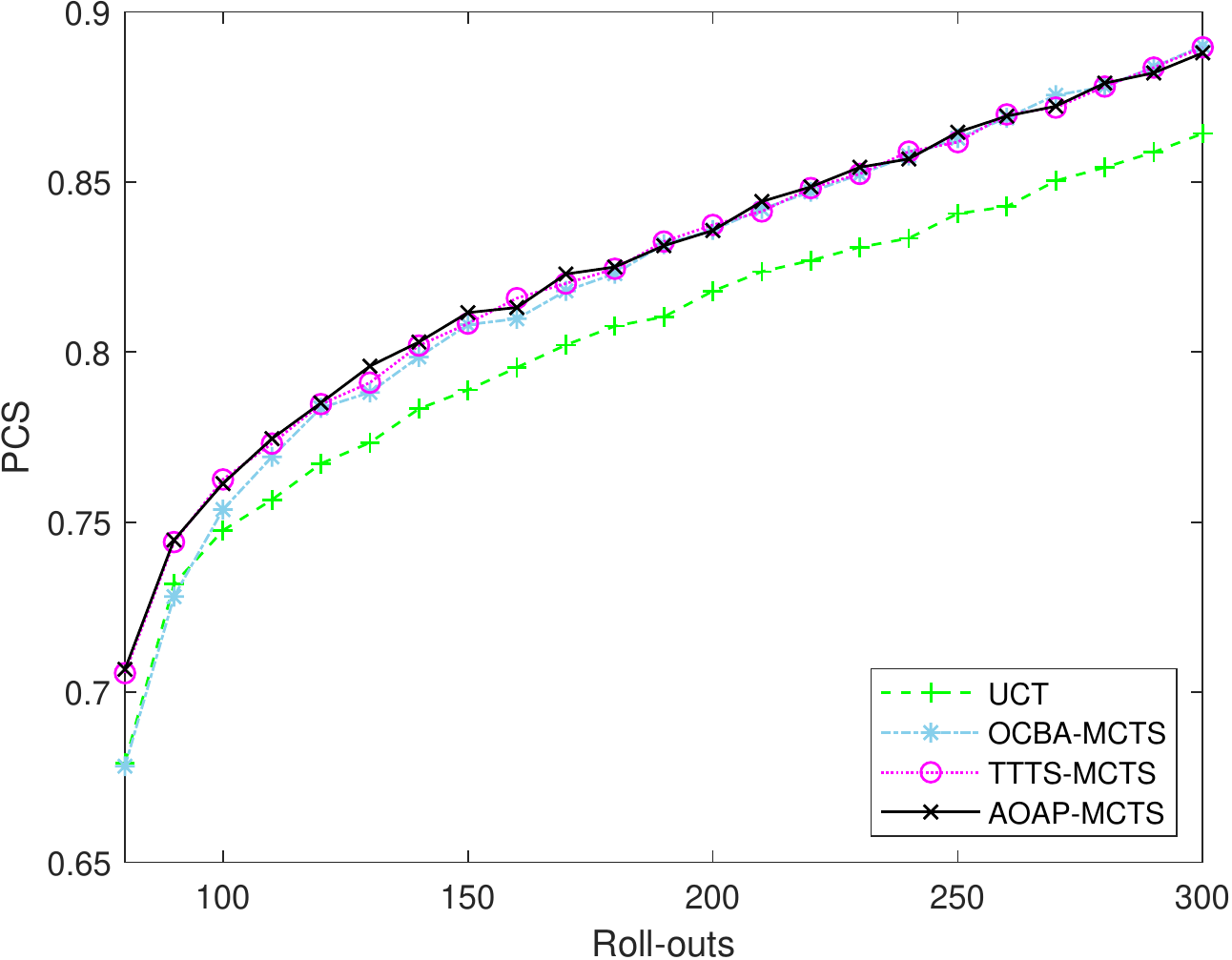}
}
\quad
\subfigure[Player 1 plays UCT in Setup 1]{
\includegraphics[width=0.43\textwidth]{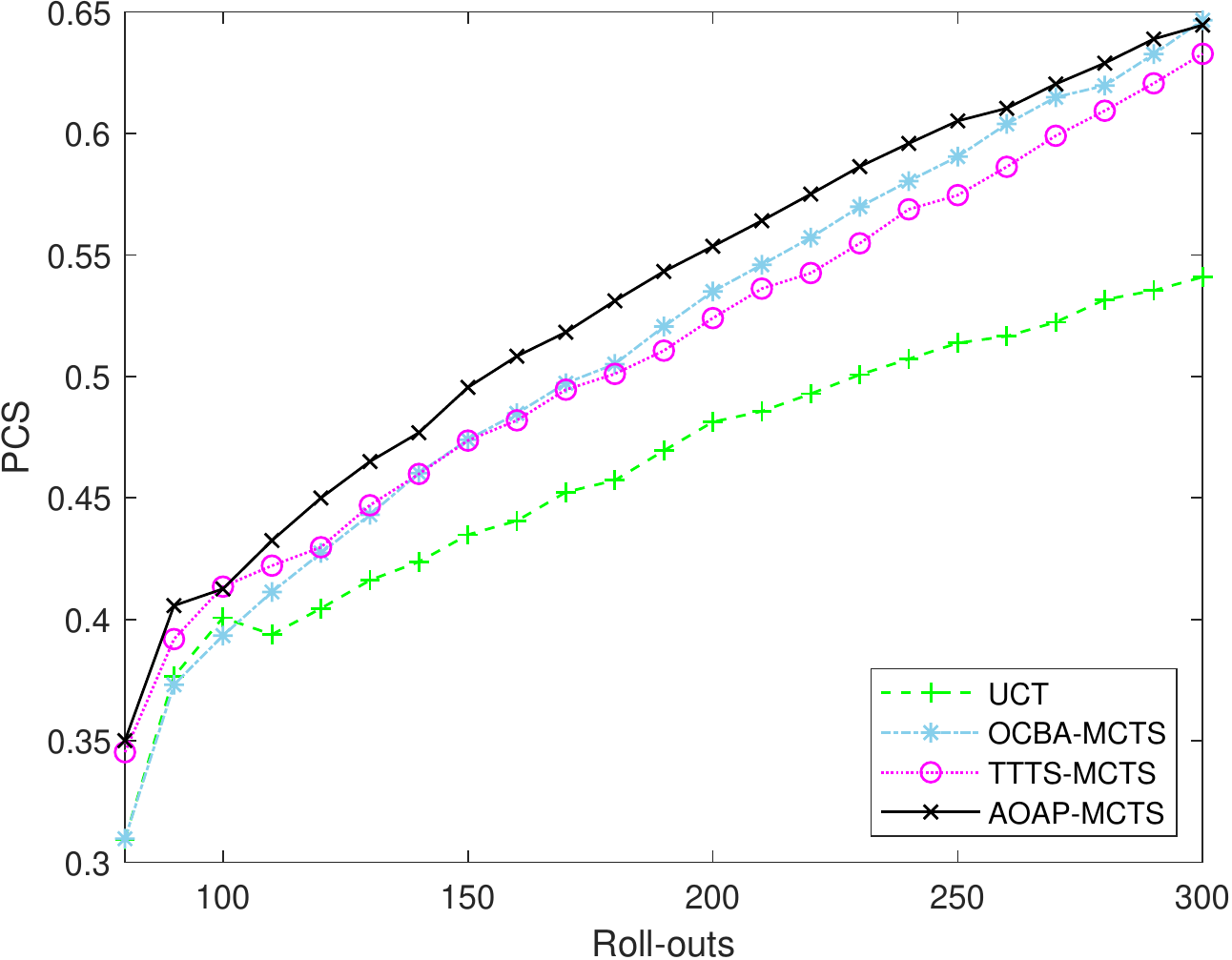}
}
\subfigure[Player 1 plays UCT in Setup 2]{
\includegraphics[width=0.43\textwidth]{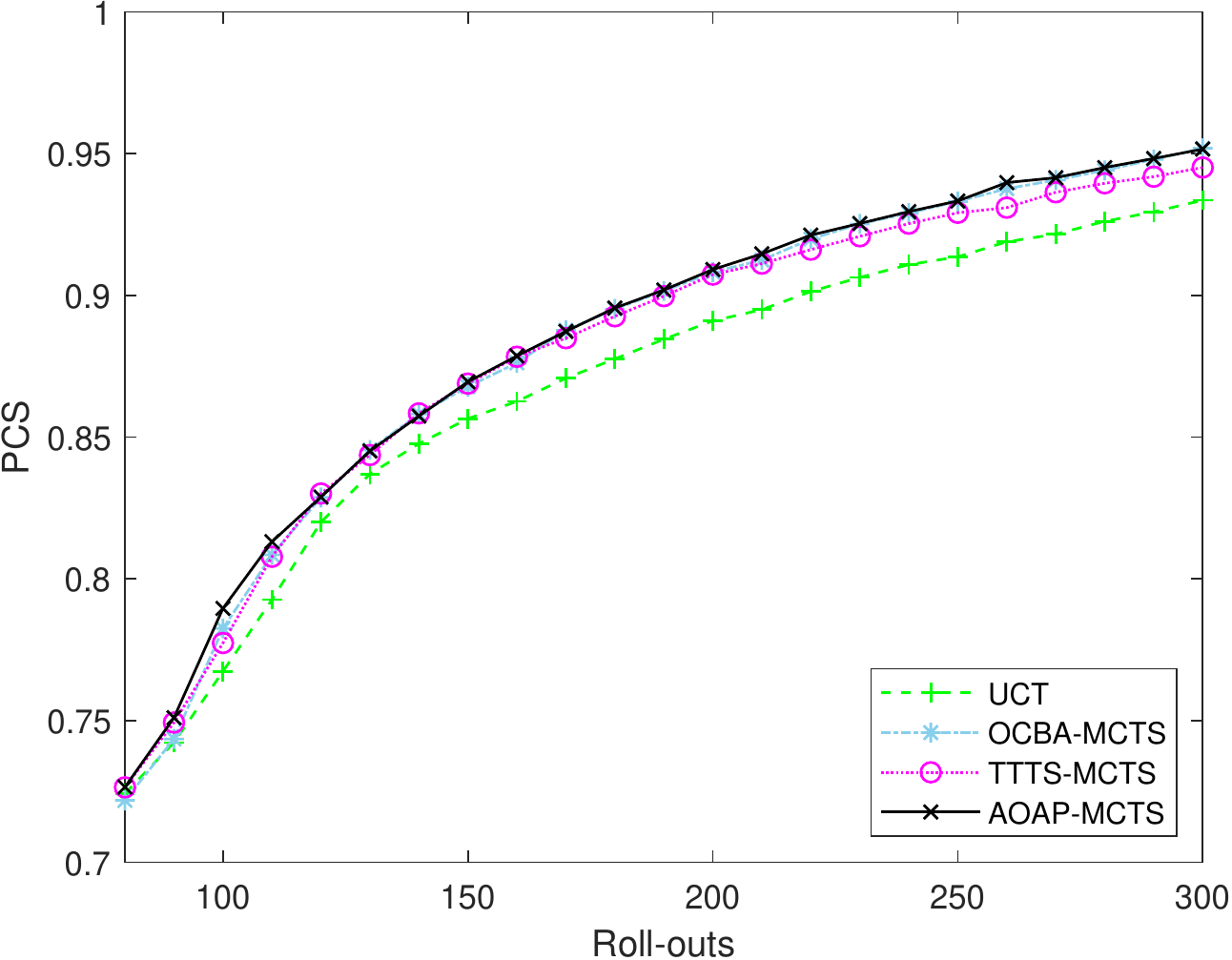}
}
\caption{Comparison of PCS of 4 tree policies in Experiment 1.1.}
\label{fig1.1res}
\end{figure}

The optimal move for Player 2 is unique in setup 1, whereas any of the four moves in the corner space is optimal for Player 2 in setup 2. The setup 2 is an easier setting since Player 2 has a 50\% chance of choosing an optimal move even if choosing randomly. At the end of the game, if Player 2 wins, the reward of terminal state is 1, and if it leads to a draw, the reward is 0.5; otherwise, the reward is 0. We consider two policies for Player 1 under both setups: one is playing randomly, i.e., with equal probability to mark any feasible space, the other is playing UCT, which chooses the move that minimizes the lower confidence bound, i.e.,
$$\widehat {\mathbf{a}}_i^{\left( t \right)} = \arg {\min\nolimits_{{\mathbf{a}_i} \in {\mathcal{A}_{{\mathbf{s}_i}}}}}\left\{{{\bar Q}_i}\left( {{\mathbf{s}_i},{\mathbf{a}_i}} \right) - {C_p}\sqrt {{{2\log {N_i}\left( {{\mathbf{s}_i}} \right)}} / {{{N_{i + 1}}\left( {{\mathbf{s}_{i + 1}}} \right)}}}\right\}~,$$
in order to minimize the reward of Player 2. We set $n_0=10$ for all policies, and set $\epsilon = 10^{-5}$ , $Q_i^{\left( 0 \right)}\left(\mathbf{s}_{i},\mathbf{a}_{i}\right) = 0$, $\sigma_i^{\left( 0 \right)}\left(\mathbf{s}_{i},\mathbf{a}_{i}\right) = 10$ for AOAP-MCTS. The PCS for the optimal move of Player 2 are estimated based on 100,000 independent macro experiments. We plot the PCS of all policies under each setup as a function of the number of roll-outs, ranging from 80 to 300. The results are shown in Figure \ref{fig1.1res}.

We can see that AOAP-MCTS performs the best among all tree policies and it has a better performance when the number of roll-outs is relatively low. The policies based on R\&S (i.e., AOAP-MCTS and OCBA-MCTS) have better performances than the policies based on MAB (i.e., UCT and TTTS-MCTS) as the number of roll-outs increases. TTTS-MCTS has a better performance than OCBA-MCTS when the number of roll-outs is low. The performances of all policies become comparable as the number of roll-outs grows. AOAP-MCTS achieves 33.2\%, 2.8\%, 19.2\% and 1.9\% better than UCT in (a)-(d) settings, respectively. The gap of policies is smaller when Player 1 plays UCT, since Player 1 has a better chance to take optimal action in this case. Although the differences between all policies in Setup 2 are not as significant as that in Setup 1, AOAP still performs the best.

\textit{Experiment 1.2: Win-draw-lose} In this experiment, we focus on the number of win, draw and lose when Player 1 plays against with Player 2. Both players play randomly or one of the four tree policies. Since opponent's policies are unknown, the player's policy is trained against a random or UCT opponent. The algorithmic constants are the same as in Experiment 1, except $Q_i^{\left( 0 \right)}\left(\mathbf{s}_{i},\mathbf{a}_{i}\right) = 1$. The number of roll-outs to determine a move at a state is set to 200. The number of win, draw and lose of Player 1 are estimated by 1000 independent rounds. The results are shown in Table \ref{table1} and Table \ref{table2}, where the trivariate vector in each blank comprises of number of win, draw and lose, respectively. The last column of each Table shows the net win of a policy, calculated by the cumulative wins minus the cumulative loses of both players.

\begin{table}[!h]
\caption{The number of win, draw and lose in Experiment 1.2, where each policy is trained against a random opponent}
\label{table1}
\centering
\begin{tabular}{c|cccccc}
\toprule
\diagbox [width=7em] {Player 1}{Player 2} & Random & UCT & OCBA-MCTS & TTTS-MCTS & AOAP-MCTS & Net Win \\
\hline
Random & (526,449,25) & (425,552,23) & (504,475,21) & (454,520,26) & (469,513,18) & -483  \\
\hline
UCT & (604,391,5) & (446,545,9) & (404,588,8) & (319,671,10) & (408,586,6) & -142  \\
\hline
OCBA-MCTS & (574,415,11) & (439,551,10) & (522,467,11) & (497,491,12) & (527,463,10) & 156 \\
\hline
TTTS-MCTS & (557,431,12) & (455,538,7) & (484,506,10) & (501,490,9) & (402,589,9) & 135 \\
\hline
AOAP-MCTS & (555,430,15) & (583,403,14) & (493,499,8) & (513,477,10) & (469,522,9) & 334 \\
\hline
\end{tabular}
\end{table}

\begin{table}[!h]
\caption{The number of win, draw and lose in Experiment 1.2, where each policy is trained against a UCT opponent}
\label{table2}
\centering
\begin{tabular}{c|cccccc}
\toprule
\diagbox [width=7em] {Player 1}{Player 2} & Random & UCT & OCBA-MCTS & TTTS-MCTS & AOAP-MCTS & Net Win \\
\hline
Random & (434,545,21) & (307,658,35) & (291,672,37) & (288,689,23) & (288,693,19) & -633  \\
\hline
UCT & (476,507,17) & (340,649,11) & (278,713,9) & (258,730,12) & (204,787,9) & -2  \\
\hline
OCBA-MCTS & (415,579,6) & (314,682,4) & (286,703,11) & (312,678,10) & (270,721,9) & 227 \\
\hline
TTTS-MCTS & (416,572,12) & (297,696,7) & (279,710,11) & (293,692,15) & (277,715,8) & 93 \\
\hline
AOAP-MCTS & (435,551,14) & (307,685,8) & (278,708,14) & (341,643,16) & (279,704,17) & 315 \\
\hline
\end{tabular}
\end{table}

From Tables \ref{table1} and \ref{table2}, we can see that the net win of AOAP-MCTS is the highest among all policies. OCBA-MCTS has a better performance than TTTS-MCTS and UCT. The net wins of AOAP-MCTS and TTTS-MCTS trained against a UCT opponent are lower than that of AOAP-MCTS and TTTS-MCTS trained against a  random opponent, showing that both policies are relatively conservative when the opponent has a better chance to take an optimal action.

\begin{figure}[!h]
\centering
\subfigure[OCBA; \textbf{AOAP}]{
\includegraphics[width=0.23\textwidth]{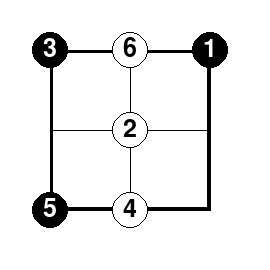}
}
\subfigure[OCBA; \textbf{UCT}]{
\includegraphics[width=0.23\textwidth]{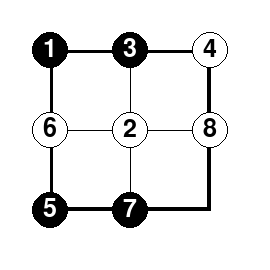}
}
\subfigure[\textbf{AOAP}; TTTS]{
\includegraphics[width=0.23\textwidth]{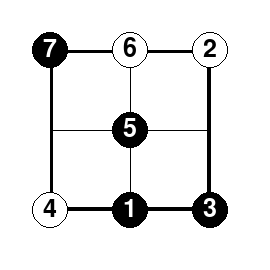}
}
\subfigure[UCT; \textbf{AOAP}]{
\includegraphics[width=0.23\textwidth]{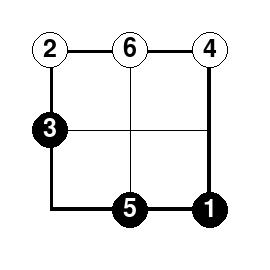}
}
\quad
\subfigure[UCT; AOAP]{
\includegraphics[width=0.23\textwidth]{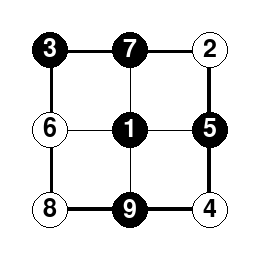}
}
\subfigure[OCBA; AOAP]{
\includegraphics[width=0.23\textwidth]{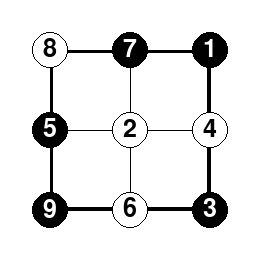}
}
\subfigure[\textbf{TTTS}; OCBA]{
\includegraphics[width=0.23\textwidth]{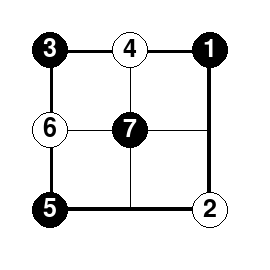}
}
\subfigure[Random; \textbf{TTTS}]{
\includegraphics[width=0.23\textwidth]{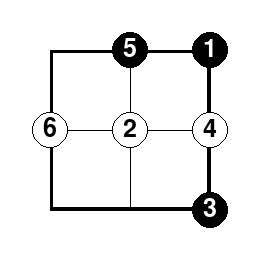}
}
\caption{Boards of Tic-tac-toe at the terminal state. The name before and after the semicolon are policies taken by Player 1 and Player 2, respectively, where we omit the `MCTS' in the name for ease of presentation. The name with the bold font is the winner.}
\label{Beha1}
\end{figure}

\textit{Experiment 1.3: Behaviors} In this experiment, we analyze the behaviors of four tree policies by observing the boards at the terminal state in games of Tic-tac-toe. Some terminal boards are shown in Figure \ref{Beha1}.

From (b), (d) and (e) in Figure \ref{Beha1}, we find that the performance of UCT does not vary much as the
game goes on. The behavior of TTTS-MCTS is similar to UCT, but it has a better performance than UCT. OCBA-MCTS tends to have a better performance at the beginning of the game, but it sometimes fails to intercept the opponent’s moves in time, leading to a failure, e.g., Figure \ref{Beha1} (a) and (b). AOAP-MCTS can aggressively intercept opponents’ moves or greedily win adaptively. Although it sometimes do not choose the optimal action at the beginning of the game, its performance becomes better as the
game goes on, e.g., Figure \ref{Beha1} (d).

\textit{Experiment 2: Gomoku} We consider a game played on a larger board, which is called Gomoku. It is played on a fifteen-by-fifteen board by two players. Players alternate turns to place a stone of their color on an empty intersection. Black plays first. The winner is the first player to form an unbroken chain of only five stones horizontally, vertically, or diagonally. We restrict the board size to eight-by-eight for ease of computation.

\begin{figure}[!h]
\centering
\includegraphics[width=0.46\textwidth]{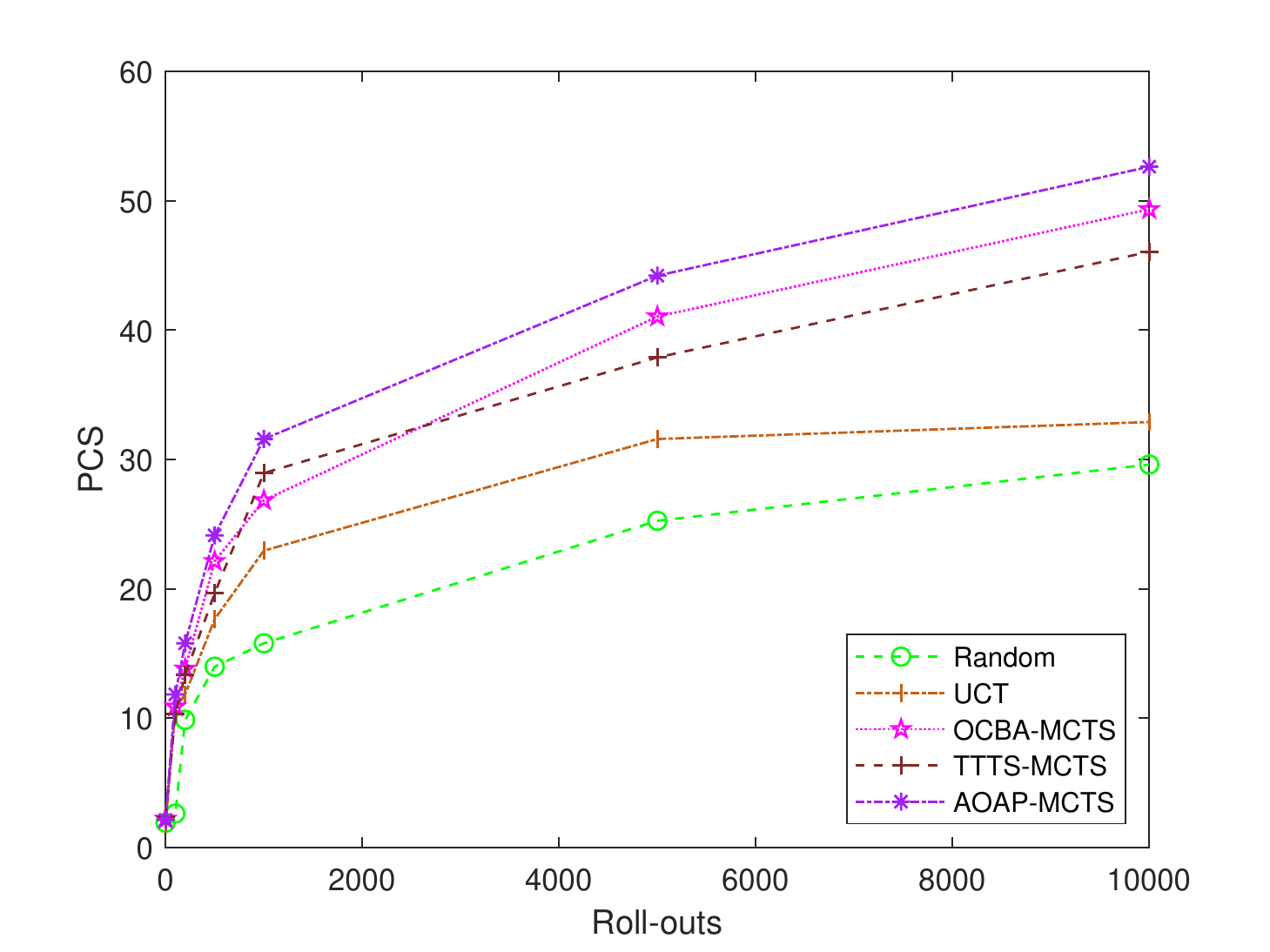}
\caption{Comparison of PCS of 5 tree policies in Experiment 2.1.}
\label{E2Z}
\end{figure}

\textit{Experiment 2.1: Precision} In this experiment, we focus on the precision of MCTS in finding the optimal move under different tree policies. The effectiveness of a policy is measured by PCS. The algorithmic constants are the same as in Experiment 1, except $Q_i^{\left( 0 \right)}\left(\mathbf{s}_{i},\mathbf{a}_{i}\right) = 1$. The true optimal move in Gomoku at a given state is more difficult to determine compared with Tic-tac-toe. In order to identify the true optimal moves, we use two random policies to play against each other, and record the change in number of win of each move using two neural networks. A move is considered to be optimal if its number of win increases more than 50\%, and the number of such a move does not exceed the half number of the Gomoku board. PCS are estimated by 100 independent states of board. Each board is estimated based on 100,000 independent macro experiments. We plot the PCS of all policies under as a function of the number of roll-outs, ranging from 0 to 10000. The results are shown in Figure \ref{E2Z}.

We can see that AOAP-MCTS performs the best among all tree policies. OCBA-MCTS has a better performance than TTTS-MCTS that is better than UCT and Random.

\textit{Experiment 2.2: Win-draw-lose} We focus on the number of win, draw and lose when Player 1 plays against with Player 2. The setups of the experiment are the same as in Experiment 1.2. The algorithmic constants are the same as in Experiment 1.2, except $(\sigma_i^{\left( 0 \right)}\left(\mathbf{s}_{i},\mathbf{a}_{i}\right) )^2 = 36$. The number of roll-outs to determine a move at a state is set to 2000. The results are shown in Table \ref{table3} and Table \ref{table4}.

\begin{table}[!h]
\caption{The number of win, draw and lose in Experiment 2.2, where each policy is trained against a random opponent}
\label{table3}
\centering
\begin{tabular}{c|cccccc}
\toprule
\diagbox [width=7em] {Player 1}{Player 2} & Random & UCT & OCBA-MCTS & TTTS-MCTS & AOAP-MCTS & Net Win \\
\hline
Random & (501,0,499) & (392,5,603) & (401,5,594) & (481,4,515) & (450,13,537) & -1549  \\
\hline
UCT & (540,4,456) & (472,14,514) & (412,4,584) & (471,11,518) & (471,0,529) & -374  \\
\hline
OCBA-MCTS & (660,7,333) & (580,10,610) & (560,5,435) & (521,12,467) & (432,14,554) & 635 \\
\hline
TTTS-MCTS & (660,11,329) & (531,5,464) & (480,14,506) & (511,0,489) & (542,1,457) & 377 \\
\hline
AOAP-MCTS & (641,0,359) & (571,13,416) & (591,3,406) & (552,3,445) & (551,2,447) & 635 \\
\hline
\end{tabular}
\end{table}

\begin{table}[!h]
\caption{The number of win, draw and lose in Experiment 2.2, where each policy is trained against a UCT opponent}
\label{table4}
\centering
\begin{tabular}{c|cccccc}
\toprule
\diagbox [width=7em] {Player 1}{Player 2} & Random & UCT & OCBA-MCTS & TTTS-MCTS & AOAP-MCTS & Net Win \\
\hline
Random & (500,7,493) & (510,9,481) & (432,49,519) & (331,4,665) & (211,7,782) & -1814  \\
\hline
UCT & (481,2,511) & (521,12,477) & (421,3,576) & (341,17,652) & (350,6,644) & -1619  \\
\hline
OCBA-MCTS & (610,11,379) & (661,5,334) & (511,7,482) & (540,5,455) & (481,3,516) & 901 \\
\hline
TTTS-MCTS & (691,7,302) & (550,3,447) & (501,16,483) & (510,3,487) & (361,9,630) & 717 \\
\hline
AOAP-MCTS & (621,19,360) & (681,8,311) & (661,9,330) & (541,2,457) & (531,6,463) & 2215 \\
\hline
\end{tabular}
\end{table}

We can see that AOAP-MCTS performs the best among all policies, and OCBA-MCTS has a better performance than TTTS-MCTS and UCT. Compared with Experiment 1.2, the advantage of AOAP-MCTS is more significant in the larger board.

\begin{figure}[!h]
\centering
\subfigure[\textbf{AOAP}; OCBA]{
\includegraphics[width=0.23\textwidth]{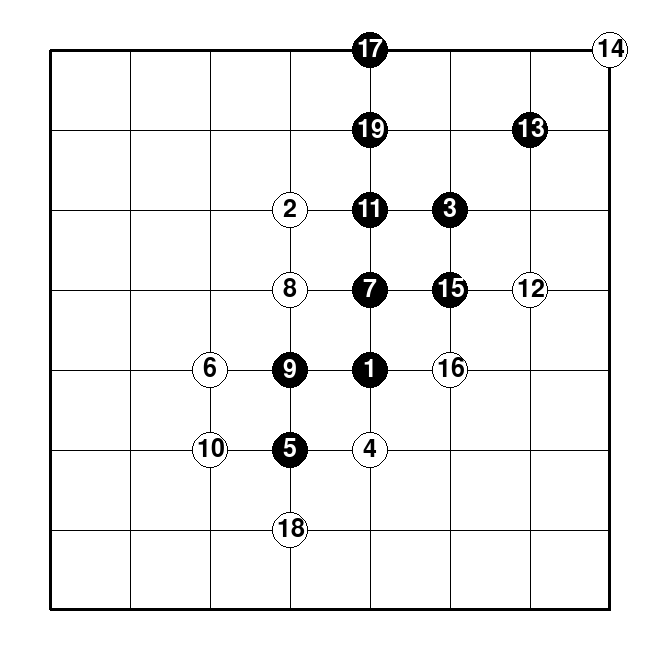}
}
\subfigure[OCBA; \textbf{UCT}]{
\includegraphics[width=0.23\textwidth]{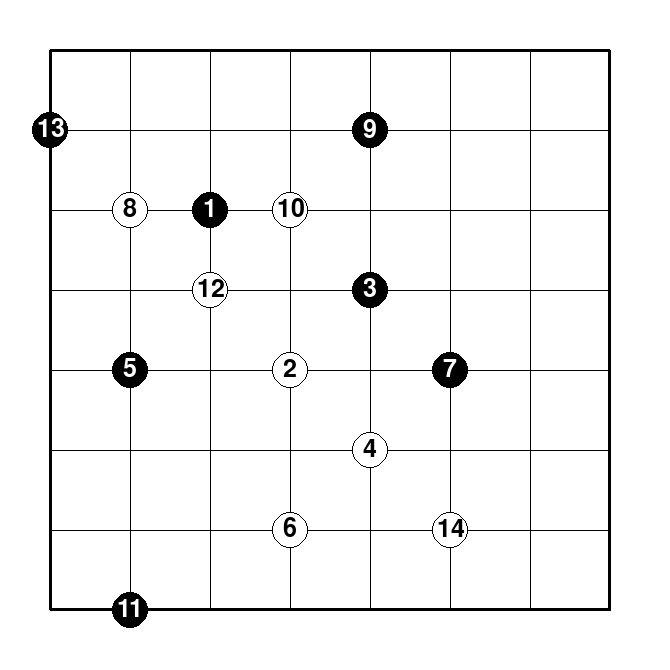}
}
\subfigure[\textbf{AOAP}; OCBA]{
\includegraphics[width=0.23\textwidth]{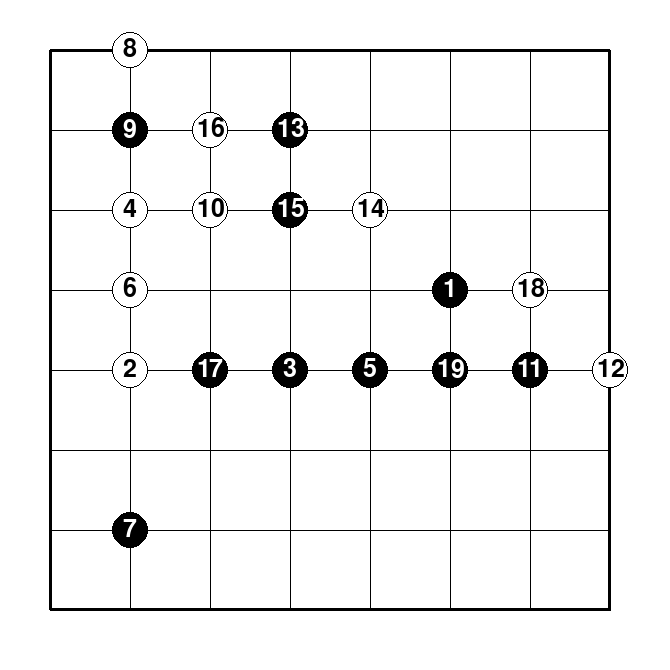}
}
\subfigure[\textbf{AOAP}; UCT]{
\includegraphics[width=0.23\textwidth]{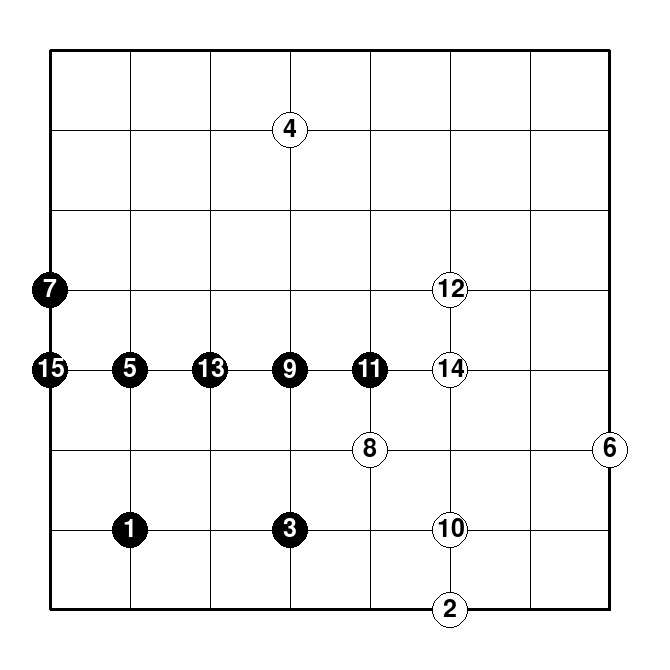}
}
\quad
\subfigure[OCBA; \textbf{AOAP}]{
\includegraphics[width=0.23\textwidth]{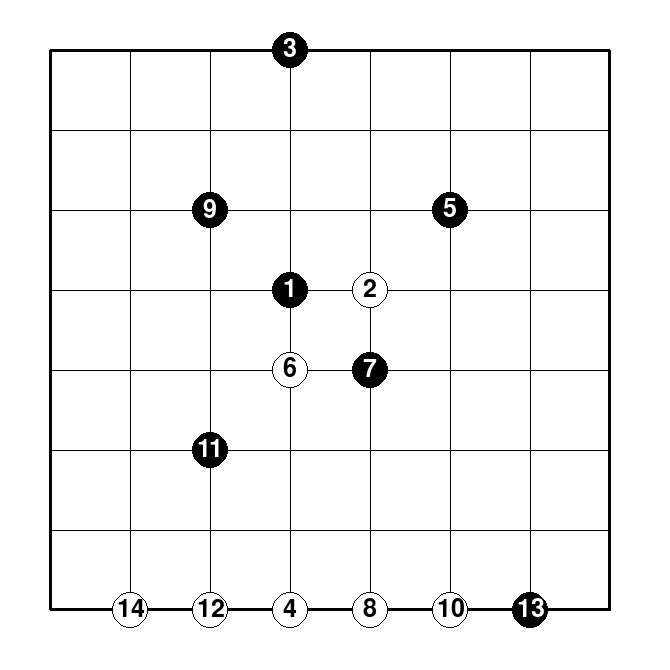}
}
\subfigure[AOAP; \textbf{AOAP}]{
\includegraphics[width=0.23\textwidth]{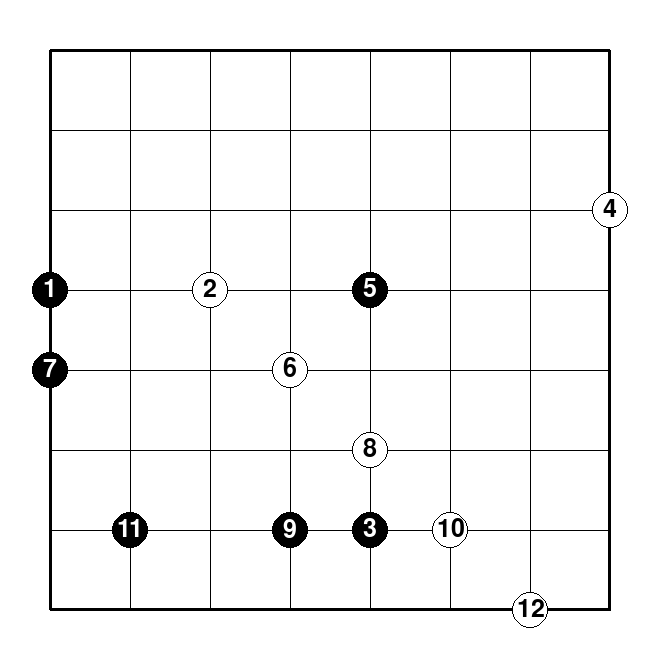}
}
\subfigure[\textbf{TTTS}; UCT]{
\includegraphics[width=0.23\textwidth]{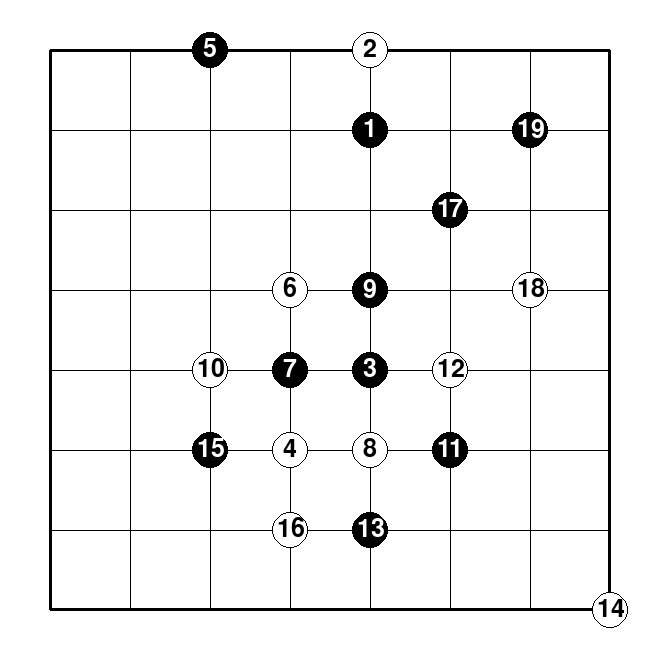}
}
\subfigure[TTTS;\textbf{AOAP}]{
\includegraphics[width=0.23\textwidth]{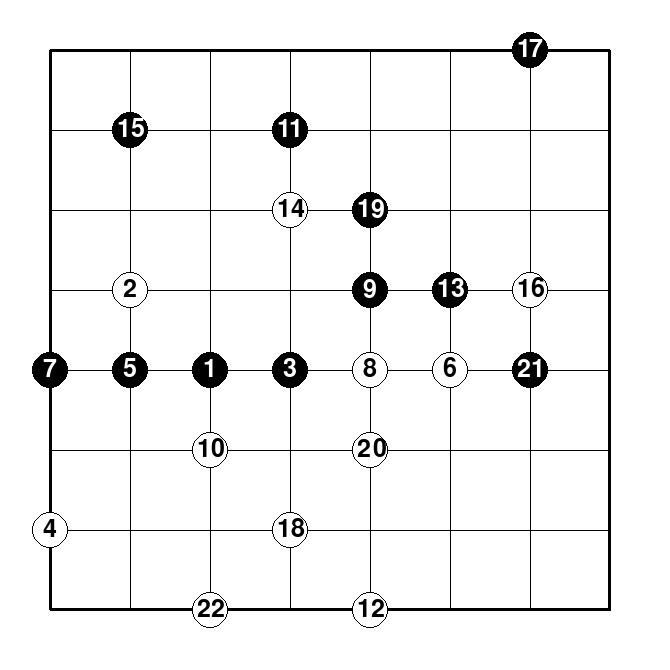}
}
\caption{Boards of Gomoku at the terminal state. The name before and after the semicolon are policies taken by Player 1 and Player 2, respectively, where we omit the `MCTS' in the name for ease of presentation. The name with the bold font is the winner.}
\label{figgrid}
\end{figure}

\textit{Experiment 2.3: Behaviors} In this experiment, we analyze the behaviors of four tree policies by observing the boards at the terminal state in games of Gomoku. Some terminal boards are shown in Figure \ref{figgrid}.

The behavior of each policy observed from Figure \ref{figgrid} is the same as in Figure \ref{fig1.1res}. UCT does not vary much as the game goes on. The behavior of TTTS-MCTS is similar to UCT, but it has a better performance than UCT. OCBA-MCTS tends to have a better performance at the beginning of the game, but it sometimes fails to intercept the opponent's moves in time, leading to a failure, e.g., (a) and (c) in Figure \ref{figgrid}. AOAP-MCTS can aggressively intercept opponents' moves or greedily win adaptively. Although it sometimes do not choose the optimal action at the beginning of the game, its performance becomes better as the game goes on.

\section{CONCLUSION}

The paper studies the tree policy for Monte Carlo Tree Search. We formulate the tree policy in MCTS as a ranking and selection problem. We propose an efficient dynamic sampling tree policy named as AOAP-MCTS, which maximizes the probability of correct selection of the best action at the root state. Numerical experiments demonstrate that AOAP-MCTS is more efficient than other tested tree policies. Future research includes the theoretical analysis of the proposed tree policy. The normal assumption of samples for $Q\left(\mathbf{s},a\right)$ deserves verification. How to guarantee sampling precision under limited computational budget could also be a future work.

\section*{Acknowledgments}

This work was supported in part by the National Natural Science Foundation of China (NSFC) under Grants 71901003 and 72022001.

\bibliographystyle{informs2014} 
\bibliography{demobib}





\end{document}